\documentclass[10pt,conference]{IEEEtran}

\usepackage{float}
\usepackage{xspace}
\usepackage{CJKutf8}
\usepackage{multirow}
\usepackage{amsmath}
\usepackage{amssymb}
\usepackage{stfloats}
\usepackage{caption}
\usepackage[hidelinks]{hyperref}
\usepackage{graphicx}
\usepackage{color}
\AtBeginDvi{\input{zhwinfonts}}

\newcommand{\WeChat}{WeChat}

\newcommand{\chstext}[1]{\begin{CJK*}{UTF8}{gkai}#1\end{CJK*}}

\newcommand{\eg}{\hbox{\emph{e.g.}}\xspace}
\newcommand{\ie}{\hbox{\emph{i.e.}}\xspace}
\newcommand{\wrt}{\hbox{\emph{w.r.t.}}\xspace}

\title{Testing Untestable Neural Machine Translation: An Industrial Case}

\author{
\IEEEauthorblockN{
Wujie Zheng\IEEEauthorrefmark{1},
Wenyu Wang\IEEEauthorrefmark{2},
Dian Liu\IEEEauthorrefmark{1},
Changrong Zhang\IEEEauthorrefmark{1},
Qinsong Zeng\IEEEauthorrefmark{1}, \\
Yuetang Deng\IEEEauthorrefmark{1},
Wei Yang\IEEEauthorrefmark{3},
Pinjia He\IEEEauthorrefmark{4},
Tao Xie\IEEEauthorrefmark{2}
}
\IEEEauthorblockA{
\IEEEauthorrefmark{1} Tencent Inc., China \\
\{wujiezheng, dianeliu, chrongzhang, qinzzeng, yuetangdeng\}@tencent.com
}
\IEEEauthorblockA{
\IEEEauthorrefmark{2} University of Illinois at Urbana-Champaign, USA \\
\{wenyu2, taoxie\}@illinois.edu
}
\IEEEauthorblockA{
\IEEEauthorrefmark{3} University of Texas at Dallas, USA \\
wei.yang@utdallas.edu
}
\IEEEauthorblockA{
\IEEEauthorrefmark{4} ETH Zurich, Switzerland \\
pinjiahe@gmail.com
}
}

\newcommand{\translationfailures}{translation failures}
\newcommand{\translationfailure}{translation failure}

\begin{document}

\maketitle

\begin{abstract}
%
Neural Machine Translation (NMT) has been widely adopted recently due to its advantages compared with the traditional Statistical Machine Translation (SMT). 
However, an NMT system still often produces \translationfailures{} due to the complexity of natural language  and sophistication in designing neural networks. 
While in-house black-box system testing based on reference translations (\ie, examples of valid translations) has been a common practice for NMT quality assurance, an increasingly critical industrial practice, named \textit{in-vivo testing},  exposes unseen types or instances of \translationfailures{} when real users are using a deployed industrial NMT system.  
To fill the gap of lacking test oracle for in-vivo testing of an NMT system, in this paper, we propose a new approach for automatically identifying  \translationfailures{}, without  requiring reference translations for a translation task; our approach can directly serve as a test oracle for in-vivo testing.
Our approach focuses on  properties of natural language translation that can be checked systematically and uses information from both the test inputs (\ie, the texts to be translated) and the test outputs (\ie, the translations under inspection) of the NMT system.
Our evaluation conducted on real-world datasets shows that our approach can effectively detect targeted property violations as \translationfailures{}. Our experiences on deploying our approach in both production and development environments of \WeChat{} (a messenger app with over one billion monthly active users) demonstrate high effectiveness of our approach along with high industry impact.

\end{abstract}


\section{Introduction} 
\label{sec:introduction}

Neural Machine Translation (NMT)~\cite{bengio2003neural,bahdanau2014neural,sutskever2014sequence} has been widely adopted over recent years due to its simple models and satisfactory effectiveness on various machine translation tasks compared with the traditional Statistical Machine Translation (SMT)~\cite{brown1993mathematics,koehn2003statistical}. 
Although NMT has shown great advantage and is becoming increasingly popular, in practice it often produces unexpected \translationfailures{} in its translations. 
For example,  Google Translate was recently reported to have various \translationfailures{}~\cite{hofstadter18:shallowness} such as producing the same translation for ``A is worse than B'' and ``B is worse than A''~\cite{google_trans_mistake_betterthan}. 
In our own empirical investigation, we also find various \translationfailures{} in translations from other highly popular providers of translation services.
Past various incidents~\cite{Okrent16:little} show that \translationfailures{}  could lead to serious consequences. For example, in 2009, a \translationfailure{} (mistranslating a catchphrase ``Assume Nothing'' to ``Do Nothing'' in various countries) caused the HSBC bank \$10 million to repair the caused damage~\cite{Okrent16:little}.

Aiming to eliminate in-field  \translationfailures{} for an NMT system, in-house quality assurance techniques such as black-box system testing commonly help NMT developers discover \translationfailures{}, subsequently enabling the developers to address these \translationfailures{}  before  the NMT system is deployed to serve the users.
In particular, the developers construct test tests from parallel corpora, which are large collections of bilingual text pairs.
Each test case corresponds to a bilingual text pair, which consists of (1) the original text (\ie, the text to be translated), being the test input, and (2) one or more reference translations, being the expected test output.
Note that reference translations are usually provided by human and are considered as examples of valid translations (\ie, samples of expected outputs from the NMT system). 
Therefore, the developers cannot simply adopt the traditional test oracle here: checking whether the NMT system's actual output (\ie, the translation under inspection) is equal to one of the reference translations. Instead, the developers typically adopt a special test oracle that (1) calculates the translation quality score (\eg, BLEU score~\cite{papineni2002bleu}) for measuring multi-granular natural-language similarity between the actual output and the reference translations, and (2) checks whether the calculated translation quality score is equal to or greater than a threshold (either a fixed or varying value), in order to determine whether the test case passes.

While black-box system testing has been shown useful for in-house quality assurance of an industrial NMT system, \textit{in-vivo testing}~\cite{murphy2009quality}, which executes tests in the production environment, is also becoming increasingly critical in industrial cases. 
In particular, by leveraging translation requests from real users in the production environment as test inputs, in-vivo testing can expose unseen types or instances of \translationfailures{} not revealed by in-house black-box system test cases, whose quantity and variety are usually much limited. 
In addition, in-vivo testing enables the developers to monitor and handle \translationfailures{}  instantly (\eg, by switching to apply a backup model or multiple models of higher quality but with higher cost) in the production environment, improving the overall translation quality on translation requests from real users.

However, during in-vivo testing for an industrial NMT system, existing test-oracle techniques cannot be applied in the production environment. First, due to the lack of samples of expected outputs (\ie, reference translations), we cannot simply apply the special test oracle that relies on the translation quality score (as widely used in black-box system testing) to in-vivo testing. Second, although there have been heuristics~\cite{pei2017deepxplore,srisakaokul2018multiple,tian2018deeptest} on speculating the expected outputs for unlabeled data on classification-oriented or prediction-oriented machine learning systems, these heuristics cannot be directly applied to testing an NMT system due to the intractability of natural languages.

To fill the gap of lacking test oracle for in-vivo testing of an industrial NMT system, in this paper, we propose a new approach for automatically identifying  \translationfailures{}; our approach can directly serve as a test oracle for in-vivo testing.
Different from using the translation quality score, our approach conducts statistical analysis on both the inputs (\ie, the original texts) and outputs (\ie, the translations under inspection) of the NMT system, aiming to directly spot out  \translationfailures{}  without samples of the expected outputs (\ie, the reference translations).
Our key insight underlying our approach is that there are some properties of natural language translation that can be checked systematically.
If there is any violation of these properties in the translation under inspection, there are likely \translationfailures{}.
In addition, we can leverage the information from the inputs of the NMT system (\ie, the original texts) to provide hints for finding \translationfailures{}, whereas such information is overlooked when the translation quality score is calculated. 
In this paper, we focus on one property that generally holds  for translations: the original text and the translation generally have one-to-one mappings in terms of their constituents, \eg, words/phrases.

In particular, our approach includes two algorithms for detecting two specific types of violations of this property, respectively: (1)  \textit{under-translation}, where some words/phrases from the original text are missing in the translation, and (2)  \textit{over-translation}, where some words/phrases from the original text are unnecessarily translated multiple times.
Based on our experience in deploying NMT models in large-scale industrial settings, many \translationfailures{}  can be reflected by these two types of violations.
For under-translation, our detection algorithm leverages the recommendation algorithm of Item-based Collaborative Filtering~\cite{sarwar2001item} to build a word/phrase translation dictionary from massive parallel corpora, and uses such algorithm for examining translations. 
Our detection algorithm additionally addresses two challenges: efficiently identifying and translating phrases from texts, and recognizing implicit word/phrase translations.
For over-translation, our detection algorithm is based on word frequencies and also considers the cases where multiple occurrences of words/phrases exist in the text to be translated.
We implement and evaluate our two algorithms as well as baseline algorithms on manually labeled datasets, which contain sets of original texts, corresponding translations, and corresponding marks of the existence for two types of violations.
Our algorithms achieve better overall performance compared with the baseline algorithms.

Our experiences on deploying our approach in both production and development environments of \WeChat{} (a messenger app with over \textbf{one billion} monthly active users~\cite{hollander18:wechat})  demonstrate that our approach is both practical and scalable. 
By using our approach for in-vivo testing in the production environment, the developers are able to collect translation tasks with unseen types or instances of \translationfailures{}  and observe the performance of deployed models in real-world usages. 
The developers are also able to handle the \translationfailures{}  instantly through switching to backup models (\eg, SMT models), improving the overall translation quality nearly effortlessly.
On top of that, our approach is able to process about \textbf{12 million} unique translation tasks every day.

In addition, not limited to only in-vivo testing, our approach  also helps enhance in-house black-box system testing during the NMT-system development.
By using our approach, the  developers manage to find outputs from the NMT system (\ie, the translations) that contain \translationfailures{}  and are previously unable to be processed by the special test oracle based on the translation quality score  (due to missing reference translations). The developers are also able to quickly locate and diagnose \translationfailures{}  when test cases fail, based on the information provided by our approach.

Finally, our approach helps build in-house testing sets containing \textbf{130,000} English and \textbf{180,000} Chinese words/phrases.
With such valuable test sets, we test multiple other highly popular  translation services and find various \translationfailures{}.  
The detected \translationfailures{}  indicate potential defects of the design, implementation, or training data in machine translation systems used by these service providers and help these providers improve their services.
All these preceding results demonstrate high effectiveness of our approach along with high industry impact.
We also build an online demonstration\footnote{http://bit.ly/2P4hEB4} for our approach on English and Chinese translation tasks.

\section{Background} 
\label{sec:background}

In this section, we introduce background of NMT quality assurance. 
We first explain why testing neural network (NN) software is a challenging task. 
Then, we discuss recent work on testing NN software. 
After that, we introduce two common types of translation-property violations that we focus on (\ie, under-translation and over-translation) with examples. 
Finally, we present BLEU score~\cite{papineni2002bleu}, one of the most widely used translation quality scores.

\subsection{Why NN Software is Hard to Test}
Although there are various studies on automatically testing traditional software, testing NN software still largely relies on manual and non-systematic strategies. 
Typically, NN software developers often conduct testing with a randomly selected testing dataset.
Because the size of the testing dataset is limited and the testing dataset is constructed in a random manner, 
these existing strategies are limited at finding and locating potential issues.  
In addition, it is inappropriate for NN software developers to directly adopt automated testing techniques proposed for traditional software for the following two main reasons.

(1) \textit{Learning instead of implementing program logics}~\cite{yang2018telemade}.
For traditional software, improving the coverage of program control-flow structures (\eg, path coverage) is the main target of automated testing techniques,
because for traditional software,  the program logics lie in its control-flow statements. 
However, the core logics of NN software are embedded in the network structure and parameters, which cannot be revealed by existing coverage metrics based on control-flow structures. 
In particular, network structure is often defined by only a few lines of code, whose coverage can easily reach 100\% without finding any potential issues. 

(2) \textit{Non-linearity.}
Constraint solving is a crucial component in automated testing techniques for traditional software.
Solvers using the Satisfiability Modulo Theory (SMT)~\cite{demoura11:SMT} have been quite successful for different demands such as generating high-coverage test inputs.
However, a neural network might consist of thousands or even millions of neurons with non-linear activation functions. 
These neurons might also be composed in a variety of ways according to different network structures. 
Thus, it is significantly more challenging to find the connections between inputs and outputs (or any other intermediate values), making constraint solving nearly infeasible there.

\subsection{Existing Testing Techniques for NN Software}
To address the aforementioned difficulties, a line of research work has been recently proposed on testing NN software, 
including whitebox testing \cite{pei2017deepxplore, ma2018deepgauge, Sun2018ASE}, metamorphic testing \cite{Zhang2018Deeproad, Dwarakanath2018ISSTA}, mutation testing \cite{Ma2018Deepmutation}, and empirical studies \cite{Zhang2018ISSTA}.
Specifically, inspired by path coverage for traditional software, Pei at al.~\cite{pei2017deepxplore} propose \textit{Neuron Coverage}, which evaluates how many neurons in the neural network under test have been activated (\ie, covered) in testing. 
This novel whitebox testing technique is further applied by Tian et al.~\cite{tian2018deeptest} to testing NN-based autonomous cars. 
Based on the core idea of neuron coverage, Ma et al.~\cite{ma2018deepgauge} further propose five fined-grained coverage criteria for whitebox testing of NN software.
Sun et al.~\cite{Sun2018ASE} propose the first concolic testing technique for NN software.
Zhang et al.~\cite{Zhang2018Deeproad} develop a GAN-based technique using the idea of metamorphic testing to synthesize test images with the same label under different weather conditions to test NN-based autonomous cars. 
Dwarakanath et al.~\cite{Dwarakanath2018ISSTA} design metamorphic rules to detect implementation faults for image classifiers (\eg, SVM and ResNet~\cite{Kaiming16ResNet}).
Ma et al.~\cite{Ma2018Deepmutation} study mutation testing strategies for both network structure and training data.

Although researchers have designed various testing techniques specialized for NN software, these techniques  consider only the neural networks whose outputs can be judged using simple equivalence oracles. 
For example, most of these techniques focus on image classification, where the correctness of NN outputs can be judged by trivially referring to image labels or voting among multiple models.
However, for NMT systems, there lacks a rigorous definition on the correctness of its output (\ie, whether the translation is valid) even when some of the expected outputs (\ie, reference translations) are provided.
As the result, it is difficult to adapt the existing testing techniques (for classification/prediction-oriented NN systems) to NMT quality assurance.

\subsection{Translation-Property Violations}
As introduced in Section~\ref{sec:introduction}, we focus on two specific types of violating the one-to-one mapping property for natural language translation: under-translation and over-translation.

\textit{Under-translation}.
For a specific translation task, if the translation misses one or more words/phrases from the text to be translated, then the translation is considered under-translated.
Table~\ref{tbl:alg_ex} shows two examples of under-translation.
In the first example, the underlined Chinese word corresponding to `mother' is missing in the English translation, leading to the wrong interpretation of what the person `she' refers to at the beginning of the second sentence.
In the second example, the underlined Chinese word corresponding to `desk' is missing in the English translation, making readers unaware of the object \textit{desk}.

\textit{Over-translation}.
For a specific translation task, if any word/phrase is unnecessarily translated multiple times, then the translation is considered over-translated. 
Table~\ref{tbl:alg_ex_overtrans} shows an example of over-translation, where the underlined Chinese phrase representing `can never be changed' appears four times in the translation, while there is only one occurrence (as indicated by italicized words) in the original sentence.
Such repetition is redundant since it does not change the meaning of the translation.

As can be seen, these two specific types of violations can directly affect the user experience and even lead to misunderstanding of the translation. 
Also according to our observation, these two specific types of violations can indicate more types of \translationfailures, including wrong names and incorrect interpretation of numbers.
In this paper, we propose two algorithms that can automatically and effectively detect these violations.

\subsection{Translation Quality Scores}
There are various translation quality scores that can be used to measure the overall quality of translations. 
BLEU (BiLingual Evaluation Understudy) score~\cite{papineni2002bleu} is one of the most widely-used quality scores for machine translation. 
BLEU score measures the similarity between the translation under inspection and the reference translation(s).
In particular, a higher BLEU score indicates that translation under inspection is closer to reference translation(s), being considered of higher translation quality.
The range of a BLEU score is $[0, 1]$, which is often presented as a percentage value (\ie, $[0,100]$). 
For example, if none of the tokens in the translation appears in any reference translation, the BLEU score is $0$.
On the contrary, if the translation is exactly the same as any of the reference translations, the BLEU score is $100$.


\section{Detection Algorithms for Under- and Over-translation} 
\label{sec:algorithms}

In this section, we introduce our detection algorithms for {under-translation} and {over-translation}. 
Our algorithms are based on the one-to-one constituent mapping property of natural language translation.
Specifically, we first build the mappings between both words and phrases from the source and destination languages on massive training translation task sets (\ie, parallel corpora).
Then, the two algorithms can leverage such mappings to examine the property on the translations under inspection.

\subsection{Building Mappings Between Bilingual Words and Phrases}
For the very beginning, based on massive training translation task sets (\ie, parallel corpora), we aim to build the mappings from each word/phrase in the source language to a list of words/phrases in the destination language, where each word/phrase in the destination language is a valid translation for the corresponding word/phrase in the source language.
There are two steps to achieve this goal: phrase identification and mapping learning.

\textbf{Phrase identification.}
The purpose of identifying phrases from texts is to handle the situations where the phrase meanings cannot be conveyed by the combination of word meanings.
The identified phrases from texts of both source and destination languages are essential to build an accurate mapping.
As a naive approach to identify phrases, we can assume that each phrase  contains up to $l_{ph}$ words, extract all 2-grams, 3-grams, ..., $l_{ph}$-grams (note that here an $n$-gram is a contiguous sequence of $n$ words) from the texts, and regard them all as candidate phrases.
However, this naive approach would probably require an enormous amount of computation due to a huge number of candidate phrases, 
whose number would be approximately 
$\sum_{i=2}^{l_{ph}}|\mathcal{W}_s|^i$ for only the source language, 
where $\mathcal{W}_s$ is the word vocabulary of the source language. 
Besides, a phrase might have different forms, adding difficulties to statistical analysis when the whole word sequences are considered.

In order to improve the efficiency make phrase identification most robust, we adopt a heuristic by only considering the representative word pairs (\ie, keyword pairs) of any phrase.
In other words, we abstract a phrase $w_1w_2\cdots{}w_n$ with an ordered (potentially not unique) word pair $\langle w_a,w_b \rangle$, where $1 \leqslant a < b \leqslant n$ and the word pair appears in different forms of the phrase.
By using the heuristic, we reduce the problem of enumerating all potential candidate phrases from enumerating all short word sequences to enumerating all word pairs in a sentence.
Specifically, assume we want to identify phrases whose keywords are at most $d_{ph}$ words away from each other, then we only need to extract each word pair $\langle w_a,w_b \rangle$ ($a,b$ denotes the word order in the same sentence) of both source and destination languages from the training dataset, where $a<b$ and $b-a \leqslant d_{ph}+1$.
For example, if we set $l_{ph}=1$, we would obtain at most 5 unique word pairs from a 4-word sentence $w_1w_2w_3w_4$: $\langle w_1,w_2 \rangle, \langle w_1,w_3 \rangle, \langle w_2,w_3 \rangle, \langle w_2,w_4 \rangle, \langle w_3,w_4 \rangle$.
Using such heuristic, we reduce the number of candidate phrases to at most $|\mathcal{W}_s|^2$ for only the source language.
We further discard any word pair with less than $c_{ph}$ occurrences to reduce false positives (\ie, the word pairs that are not keyword pairs of any phrase).
Additionally, \textit{mapping learning} (as introduced later) also helps identify real phrases.

\textbf{Mapping learning.}
We employ the recommendation algorithm of Item-based Collaborative Filtering~\cite{sarwar2001item} to build the word/phrase translation mappings from massive training translation task sets (\ie, parallel corpora).
The recommendation algorithm of Item-based Collaborative Filtering was originally used in scenarios such as product and movie recommendation~\cite{linden2003amazon,davidson2010youtube,adomavicius2005toward}.
The algorithm predicts an user's prospective interested items (\eg, movies) by looking at the user's rating history and searching the items that are similar to the user's highly-rated items.
The algorithm's key insight of finding similar items is that similar items should have similar ratings from a lot of users.
Thus, it determines the similarity between two items by looking at the similarity between their ratings.
To reduce our mapping problem to the recommendation problem, we can view each training translation task as a user, each word/phrase appearing the original text (of the source language) or reference translation (of the destination language) as an item with a positive rating, and all other words/phrases as items with negative ratings.
In this way, we adapt the recommendation algorithm to uncover the connections between words/phrases in the source and destination languages, \ie, the word/phrase translations. 
Specifically, we define the user rating matrix $M$ as follows:

\vspace{-2ex}
\begin{equation*}
	M_{k,w} = 
	\begin{cases}
		1 & \text{if } w \text{ appears in } P_s^k \text{ or } P_d^k \\
		0 & \text{otherwise}
	\end{cases}
\end{equation*}
\vspace{-2ex}

\noindent where $w$ is a word/phrase, $P_s^k$ and $P_d^k$ denote the original text and the reference translation in the training translation task numbered $k$ correspondingly.
Then, we calculate the relevance/similarity between $w_s$ (a word/phrase in the source language) and $w_d$ (a word/phrase in the destination language) using the Cosine similarity~\cite{dangeti2017statistics}:

\vspace{-3ex}
\begin{align*}
	Rel(w_s, w_d) &= \frac{\overrightarrow{M_{\cdot,w_s}} \cdot \overrightarrow{M_{\cdot,w_d}}}{||\overrightarrow{M_{\cdot,w_s}}||_2 \cdot ||\overrightarrow{M_{\cdot,w_d}}||_2} \\
	&= \frac{\sum_k M_{k,w_s}M_{k,w_d}}{\sqrt{\sum_k M_{k,w_s}^2}\sqrt{\sum_k M_{k,w_d}^2}}
\end{align*}

\begin{table}[t]
\centering
\small
\caption{Example training translation tasks}
\label{tbl:ibcf_ex_corpus}
\begin{tabular}{|c|l|l|}
\hline
 & English (original) & Chinese (translated)
\\ \hline
1 & I love my family. & \chstext{我爱我的家人。} \\ \hline
2 & I have a lovely son. & \chstext{我有一个可爱的儿子。} \\ \hline
3 & I have worked for seven years. & \chstext{我工作七年了。} \\ \hline
4 & They have a big house. & \chstext{他们有一个大房子。} \\ \hline
\end{tabular}
\end{table}

\begin{table}[t]
\centering
\small
\caption{User rating matrix $M$ corresponding to Table~\ref{tbl:ibcf_ex_corpus}}
\label{tbl:ibcf_ex_mat}
\begin{tabular}{|c|ccccccccc|}
\hline
& I & love & have & $\cdots$ & \chstext{我} & \chstext{爱} & \chstext{有} & $\cdots$ & \chstext{。} \\
\hline
1 & 1 & 1 & 0 & $\cdots$ & 1 & 1 & 0 & $\cdots$ & 1 \\ \hline
2 & 1 & 0 & 1 & $\cdots$ & 1 & 0 & 1 & $\cdots$ & 1 \\ \hline
3 & 1 & 0 & 1 & $\cdots$ & 1 & 0 & 0 & $\cdots$ & 1 \\ \hline
4 & 0 & 0 & 1 & $\cdots$ & 0 & 0 & 1 & $\cdots$ & 1 \\ \hline
\end{tabular}
\vspace{-4ex}
\end{table}

Tables~\ref{tbl:ibcf_ex_mat} shows an instance of $M$ for the translation task in Table~\ref{tbl:ibcf_ex_corpus}.
We can calculate $Rel(\text{I},\text{\chstext{我}})=([1,1,1,0] \cdot [1,1,1,0])/(||[1,1,1,0]|| \cdot ||[1,1,1,0]||)=1$, indicating that we speculate `\chstext{我}' to be the translation of `I'.
In fact, such translation is exactly the correct Chinese translation of `I'.

Based on such relevance between words/phrases in the source and destination languages, we build the mappings directly.
Specifically, for each word $w_s$ in the source language, we consider a total of $c_{tr}$ words in the destination language with the highest relevance to $w_s$ to be valid translations of $w_s$, where $c_{tr}$ is a predefined threshold value.
Such technique has two advantages compared with using generic translation dictionaries.
First, the words in an existing generic translation dictionary may be limited. 
If a word is not in the existing dictionary, then the mapping cannot be constructed.
Second, the word translations included in the generic dictionary might also be too limited, too generic, or outdated with respect to various translation tasks, causing the algorithms to miss correct translations.

\subsection{Detection Algorithm for Under-translation}
\label{sec:under_trans}
With the word/phrase translation mappings derived from the previous steps, under-translation detection can now be achieved by checking the existence of word/phrase translations in the whole translation text.
Specifically, for each word/phrase in the original text, we obtain the list of its corresponding word/phrase translations, and check if any of these word/phrase translations appear in the whole translation text.
We use two real-world examples to illustrate this process. 
Table~\ref{tbl:alg_ex} shows two example translations with under-translation violations.
The underlined Chinese words (corresponding to `mother' and `desk' in English) are missing in the English translation.
When our algorithm detects potential violations related to these two words, two translation lists are available for each of them.
Specifically, our algorithm produces the contents in Table~\ref{tbl:alg_ex_trans}, which is part of the mappings constructed by the previous steps.
In the table, `origin' indicates the words to be translated and `trans $k$' denotes the $k$-th most relevant translation.
Our algorithm goes through each translation for both Chinese words and checks whether it appears in the translation being examined.
Then, the algorithm finds that for both Chinese words, none of translations in Table~\ref{tbl:alg_ex_trans} appears in the translation being examined (as shown in Table~\ref{tbl:alg_ex}).
The algorithm subsequently marks both translations with potential under-translation violations.

However, there might be some words/phrases of the source language whose translations usually do not show up in the translations of the whole sentences, \ie, some words/phrases might incur implicit translations.
Examples include \textit{the} and \textit{to} in English.
The techniques specified earlier can likely produce numerous false positives due to such phenomenon.
We introduce \textit{error rate filtering} to address such limitation.
Specifically, we calculate the {error rate} $e_{w_s}$ for each word $w_s$ of the source language (similar for any phrase) on the training dataset using $e_{w_s} = {N_{w_s}^e}/{N_{w_s}}$, 
where $N_{w_s}^e$ denotes the number of sentence pairs that are considered under-translated on $w_s$, 
and $N_{w_s}$ is the number of translation sentence pairs involving $w_s$.
Then, we disregard any under-translation caused by missing $w_s$ on real translation results if we find that $e_{w_s} > e_{th}$, where $e_{th}$ is a predefined threshold value that controls the tolerance of such error-prone $w_s$.
Table~\ref{tbl:alg_ex} shows the technique being used in two example translations, where `error\_rate' shows the error rates for two words in the source language.
The algorithm also checks whether the error rates are beyond $e_{th}$ (usually set to be $0.2$ in our production environment) for both Chinese words before marking them with potential under-translation violations. 
In the examples, the algorithm finds that both error rates are acceptable and confirm the marks.
On the contrary, if the algorithm finds that the error rate on one word/phrase is too high, it will skip the word/phrase and continue checking other words/phrases.

\begin{table*}[t]
\centering
\small
\caption{Example translations with under-translation violations}
\label{tbl:alg_ex}
\begin{tabular}{|p{3cm}|p{4cm}|p{5cm}|}
\hline
Chinese (original) & English (translated) & English (reference)
\\ \hline
\chstext{三姑给你的红包给你\underline{妈妈}了，她见了你会给你的。}
&
Third Aunt gave you a red 

envelope. She'll give it to you 

when she sees you.
&
Third Aunt gave your red envelope

to your \textit{mother}. She'll give it to you 

when she sees you.
\\ \hline
\chstext{放寒假一个月宿舍地上\underline{桌}上床上全是灰...都不知道该从哪里开始收拾..}
&
Winter vacation a month 

dormitory on the floor of the bed 

is all gray ... I don't know 

where to start .. 
&
After a month's winter vacation,

dust is everywhere on the floor, the

\textit{desk}, and the bed in the dormitory...

I don't know where to start cleaning...
\\ \hline
\end{tabular}
\vspace{-2ex}
\end{table*}

\begin{table}[t]
\centering
\small
\caption{Example translation lists for two under-translated words in Table~\ref{tbl:alg_ex}}
\label{tbl:alg_ex_trans}
\begin{tabular}{|c|l|l|}
\hline
origin	&	\chstext{妈妈} & \chstext{桌} \\ \hline
error\_rate	&	0.04116 & 0.07293 \\ \hline
trans 1	&	mother		&	table		\\
trans 2	&	mom			&	desk		\\
trans 3	&	mum			&	papers		\\
trans 4	&	mama		&	tables		\\
trans 5	&	mommy		&	coffee		\\
trans 6	&	moms		&	placed		\\
trans 7	&	mothers		&	piled		\\
trans 8	&	mummy		&	put			\\
trans 9	&	my			&	breakfast	\\
trans 10	&	her			&	laid		\\ \hline
\end{tabular}
\end{table}

\subsection{Detection Algorithm for Over-translation}
Our over-translation detection algorithm is based on frequency of words appearing in the translation. 
Specifically, we count the occurrences for each word appearing in the translation under inspection, 
and mark the word as over-translated if more than one occurrence is found and the occurrences are near to each other. 
In addition, our algorithm includes techniques to address two main challenges being faced. 

First, particles such as \textit{have}, \textit{the}, and \textit{to} in English tend to have multiple occurrences and no actual meaning in many well-formed expressions.
These words would confuse over-translation detection and cause false positives.
To alleviate this challenge, we remove all such common words (\ie, stop words) from the translation under inspection before conducting over-translation detection.

Second, some words/phrases might have multiple occurrences in the original text, and they are probably also supposed to appear multiple times in the translation under inspection.
Aiming to differentiate such situations from those with real over-translation violations, our algorithm includes a technique to estimate the number of words/phrases (in the original text) that are translated to each word $w_d$ in the translation under inspection,  and compare such number with the number of occurrences of $w_d$.
Specifically, for each word $w_d$ in the translation under inspection, we use the word/phrase translation mappings introduced by the previous steps in the other direction to find out a set of words/phrases (of the source language) that can be all translated to $w_d$ (\ie, for each word $w_s$ in the set, $w_d$ is its translation according to the dictionary), and count the number of words/phrases in the original text (denoted as $count_s(w_d)$) that fall in the set.
Let us denote the number of occurrences of $w_d$ in the translation under inspection as $count(w_d)$.
Finally, we can consider a $w_d$ to be over-translated if we find $count_s(w_d) < count(w_d)$, indicating that we find more occurrences of $w_d$ than there should be, \ie, redundant translations.

\begin{table}[t]
\centering
\small
\caption{Example translation with over-translation violation}
\label{tbl:alg_ex_overtrans}
\begin{tabular}{|p{3.7cm}|p{4.12cm}|}
\hline
English (original) & Chinese (translated)
\\ \hline
U have to admit that some-

thing \textit{can never be changed}, 

live with it or break with it!
&
\chstext{你必须承认，有些东西是永

远无法改变的，\underline{无法改变的}，

\underline{无法改变的}，\underline{无法改变的}！}
\\ \hline
\end{tabular}
\vspace{-4ex}
\end{table}

Table~\ref{tbl:alg_ex_overtrans} shows an example with an over-translation violation to illustrate our algorithm.
The Chinese translation contains 3 more repetitions for the translation of ``can never be changed''.
Our algorithm first discovers that there are 4 instances of ``\chstext{无法}'' and ``\chstext{改变}''.
Then, the algorithm looks up the word/phrase translation mappings to find whether there is any word/phrase in the original text with such translations. 
The algorithm finds that ``can never'' and ``changed'' have such translations.
Then it also finds that these two English words/phrases appear only once in the original text, fewer than the occurrences in the translation under inspection.
Finally our algorithm marks the translation with an over-translation violation.

\section{Evaluation} 
\label{sec:evaluations}

In this section, we present our evaluation on assessing our proposed approach.
Specifically, we try to answer the following two research questions (RQs):

\textit{RQ1}: How accurate are the two proposed violation detection algorithms (included in our approach) in testing benchmark datasets consisting of real-world translation tasks?

\textit{RQ2}: How useful is our approach in assisting our NMT system improvement during both in-vivo testing and in-house testing?

\textit{RQ3}: How effective is our approach on testing NMT systems from other public service providers?

We next discuss the evaluation setup (Section~\ref{sec:setup}) and present the accuracy comparison and analysis results (Section~\ref{sec:comparison}) \wrt baseline algorithms, for RQ1.
We then share our experience of deploying our proposed approach in both the production and development environment of \WeChat, a messenger app with over one billion of monthly active users (Section~\ref{sec:deploy}), for RQ2.
We also demonstrate the effectiveness of our approach on testing NMT systems from other public service providers by showing the instances of detected \translationfailures{} in the translations from those systems (Section~\ref{sec:appl_rw_tasks}), for RQ3.

\subsection{Evaluation Setup}
\label{sec:setup}
We implement our proposed detection algorithms (denoted as `proposed') from scratch, and evaluate our proposed algorithms.
We also evaluate baseline algorithms on both under- and over-translation detection for comparison with our proposed algorithms:

\textit{Algorithms based on primitive dictionary looking-up} (denoted as `std-dict').
For under-translation, we look up each word/phrase of the text to be translated in an existing generic translation dictionary and check whether any translation of the word appears in the translation.
For over-translation, we replace the learned word/phrase translation mappings in our algorithm with the existing generic translation dictionary.
We use the software \textit{StarDict}~\cite{StarDict} to provide existing generic translation dictionaries and implement such baseline algorithm by ourselves.

\textit{Algorithms based on word alignment}~\cite{koehn2009statistical,vogel1996hmm} (named as `word-align') from the traditional SMT models.
For under-translation, we let the word-alignment model provide a list of candidate translations for each word of the text to be translated, and check whether any translation of the word appears in the translation.
For over-translation, we replace the learned word/phrase translation mappings in our algorithm with the candidate translation lists provided by the word-alignment model.
Note that the candidate translation list is also derived by choosing the translations with top $c_{tr}$ alignment probabilities.
We use the software \textit{fast\_align}~\cite{dyer2013simple} with default settings for the word-alignment model implementation.

We build the word/phrase translation mappings (for our algorithms) and train the word-alignment model (for the baseline algorithm) with 16 million sentence pairs crawled from various sources (\eg, example word/phrase usages in dictionaries) on the Internet.
We evaluate all the algorithms on datasets that are randomly sampled from larger benchmark datasets (crawled online and independent of training datasets) and are manually labeled by us.
Each dataset contains a number of sentence pairs, each of which consists of the sentence to be translated (\ie, the input to our NMT system) and the translation under inspection (\ie, the corresponding output of our NMT system).
Each sentence pair is also marked with the existence of two types of violations in the translation identified by manual inspection.
Table~\ref{tbl:eval_dataset} shows the overview of the evaluation datasets, where $\#_{all}$ denotes the total number of sentence pairs,
 $\#_{ws}$ denotes the total number of words of the source language, 
 and  $\#_U$ and $\#_O$ denote the numbers of sentence pairs flagged with under- and over-translation violations, respectively.
`en-cn' and `cn-en' indicate translating from English to Chinese and from Chinese to English, respectively.

\begin{table}[t]
\centering
\small
\caption{Overview of evaluation datasets}
\label{tbl:eval_dataset}
\begin{tabular}{|c|c|c|r|r|r|r|} \hline
Name & Lang & Type & $\#_{all}$ & $\#_{ws}$ & $\#_U$ & $\#_O$ \\ \hline
ench\_news & en-cn & News & 200 & 7497 & 54 & 4 \\ \hline
chen\_news & cn-en & News & 200 & 7418 & 31 & 8 \\ \hline
ench\_oral & en-cn & Oral & 300 & 3237 & 42 & 1 \\ \hline
chen\_oral & cn-en & Oral & 300 & 2918 & 37 & 5 \\ \hline
\end{tabular}
\vspace{-3ex}
\end{table}

For any algorithm on any dataset, let $\mathcal{S}$ be the manually labeled sentence pair sets containing a specific violation type (\ie, ground truth), 
and let $\mathcal{S}'$ be the sentence pair sets identified by the algorithm detecting the specific violation type, 
then \textit{precision}, \textit{recall}, and \textit{F-measure} can be calculated as:

\vspace{-3ex}
\begin{gather*}
	\text{precision} = \frac{|\mathcal{S} \cap \mathcal{S}'|}{|\mathcal{S}'|}
	\quad
	\text{recall} = \frac{|\mathcal{S} \cap \mathcal{S}'|}{|\mathcal{S}|}
\\
	\text{F-measure} = 2 \cdot \frac{\text{precision} \cdot \text{recall}}{\text{precision} + \text{recall}}
\end{gather*}

\subsection{Effectiveness Comparison and Analysis}
\label{sec:comparison}

\textbf{General comparison.}
For the algorithms under comparison, Table~\ref{tbl:eval_results} shows the result summary of under-translation detection, and Table~\ref{tbl:eval_results_overtrans} shows the result summary of over-translation detection.
The precision, recall, and F-measure are abbreviated as `P', `R', and `F', respectively, in both tables. 
In Table~\ref{tbl:eval_results}, `Count' indicates the number of under-translated words identified by the three algorithms under comparison. 
We set $c_{tr}=10$ and $c_{ph}=10$ for all the experiments, $e_{th}=0.15$ for experiments on the `ench\_news' and `chen\_news' datasets, and $e_{th}=0.25$ for experiments on the `ench\_oral' and `chen\_oral' datasets.
We turn on phrase identification for `en-cn' tasks while keeping it off for `cn-en' tasks due to the fact that most Chinese words are already phrases after word segmentation.
As shown in Tables~\ref{tbl:eval_results} and~\ref{tbl:eval_results_overtrans}, our proposed algorithm achieves the highest F-measures on all the datasets, compared with the two baseline algorithms (`std-dict' and `word-align'). 
Such result indicates the effectiveness of our proposed algorithm.

For under-translation detection, as shown in Table~\ref{tbl:eval_results}, the higher effectiveness of our proposed algorithm mainly comes from higher precisions.
Our algorithm's precisions are up to about $2.6\times$ of those by the baseline algorithms (`std-dict' and `word-align'). 
Such result indicates the high accuracy of our combined techniques for under-translation detection.
We also notice that precisions achieved by `word-align' are higher than those by `std-dict'.
Such result shows the necessity of word/phrase translation mapping learning, which helps avoid too generic or limited word/phrase translations.
Meanwhile, we find that our algorithm has relatively lower recalls, potentially as the cost of achieving higher accuracy.
On the contrary, `std-dict' achieves the highest recalls on all the datasets.
However, such scores are accompanied by a large number of false positives, making it difficult to leverage the detection results.

\begin{table*}[tp]
\centering
\small
\caption{Results of under-translation detection on the evaluation datasets}
\label{tbl:eval_results}
\begin{tabular}{|c|r|r|r|r|r|r|r|r|r|r|r|r|r|r|r|r|r|r|r|r|r|} \cline{2-13}
\multicolumn{1}{c|}{} & \multicolumn{4}{c|}{proposed} & \multicolumn{4}{c|}{std-dict} & \multicolumn{4}{c|}{word-align} \\ \cline{2-13}
\multicolumn{1}{c|}{} & P & R & F & Count & P & R & F & Count & P & R & F & Count \\ \hline
ench\_news & \textbf{0.51} & 0.69 & \textbf{0.59} & 113 & 0.28 & \textbf{1.00} & 0.43 & 1853 & 0.35 & 0.85 & 0.49 & 250 \\ \hline
chen\_news & \textbf{0.43} & 0.65 & \textbf{0.52} &  50 & 0.16 & \textbf{1.00} & 0.27 & 1827 & 0.18 & \textbf{1.00} & 0.30 & 523 \\ \hline
ench\_oral & \textbf{0.52} & 0.40 & \textbf{0.45} &  32 & 0.15 & \textbf{1.00} & 0.26 &  689 & 0.20 & 0.79 & 0.32 & 309 \\ \hline
chen\_oral & \textbf{0.30} & 0.49 & \textbf{0.37} &  70 & 0.12 & \textbf{0.97} & 0.22 &  895 & 0.14 & 0.73 & 0.24 & 259 \\ \hline
\end{tabular}
\end{table*}

For over-translation detection, as shown in Table~\ref{tbl:eval_results_overtrans}, the higher effectiveness of our proposed algorithm also mainly comes from higher precisions.
However, different from the results of under-translation detection, rankings of recalls vary across the datasets.
Such result might be due to a relatively small number of over-translation instances in the datasets.

\begin{table*}[tp]
\centering
\small
\caption{Results of over-translation detection on the evaluation datasets}
\label{tbl:eval_results_overtrans}
\begin{tabular}{|c|r|r|r|r|r|r|r|r|r|r|r|r|r|r|r|r|r|r|} \cline{2-10}
\multicolumn{1}{c|}{} & \multicolumn{3}{c|}{proposed} & \multicolumn{3}{c|}{std-dict} & \multicolumn{3}{c|}{word-align} \\ \cline{2-10}
\multicolumn{1}{c|}{} & P & R & F & P & R & F & P & R & F \\ \hline
ench\_news & \textbf{0.38} & 0.75 & \textbf{0.50} & 0.13 & 0.50 & 0.20 & 0.24 & \textbf{1.00} & 0.38 \\ \hline
chen\_news & \textbf{0.73} & \textbf{1.00} & \textbf{0.84} & 0.13 & \textbf{1.00} & 0.23 & 0.40 & \textbf{1.00} & 0.57 \\ \hline
ench\_oral & \textbf{0.33} & \textbf{1.00} & \textbf{0.50} & 0.00 & 0.00 & 0.00 & 0.14 & \textbf{1.00} & 0.25 \\ \hline
chen\_oral & \textbf{0.80} & 0.80 & \textbf{0.80} & 0.38 & \textbf{1.00} & 0.56 & 0.28 & \textbf{1.00} & 0.43 \\ \hline
\end{tabular}
\vspace{-4ex}
\end{table*}

\begin{figure}[t]
	\centering
	\includegraphics[scale=0.65]{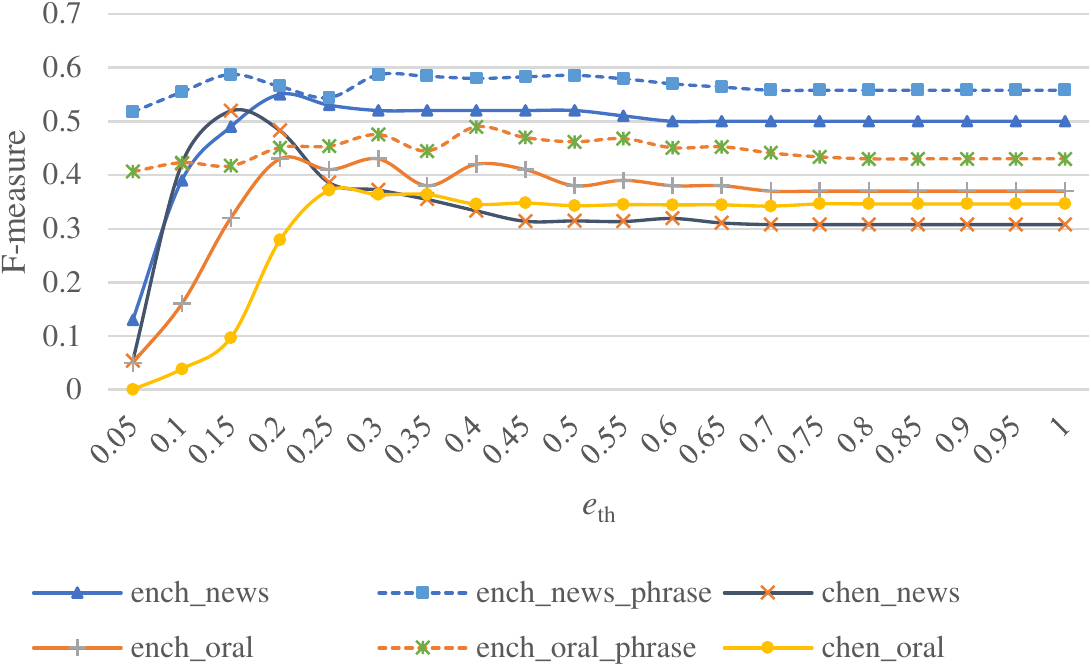}
	\caption{Trends of F-measures of the detection results on all datasets with different $e_{th}$ values}
	\label{fig:eth_vs_f1}
\vspace{-4ex}
\end{figure}

\textbf{Analysis of Error Rate Thresholds $e_{th}$.}
Changing the value of $e_{th}$ might influence the detection results.
Thus, we further investigate how different threshold values would affect the overall effectiveness of our algorithm.
As stated in Section \ref{sec:under_trans}, $e_{th}$ controls the tolerance of error-prone words in the proposed under-translation detection algorithm.
We also know from the definition that $e_{th} \in (0, 1]$.
Aiming to understand the influence of $e_{th}$ on the algorithm effectiveness, we run the under-translation algorithm on all the datasets with $e_{th}$ set to $\{0.05, 0.1, \cdots, 0.95, 1\}$.
Figure~\ref{fig:eth_vs_f1} shows the trends of F-measure of the detection results under different $e_{th}$ values.
Note that we run the algorithm on `en-ch' datasets both when phrase identification is enabled and disabled (corresponding to `ench\_news', `ench\_news\_phrase', `ench\_oral', and `ench\_oral\_phrase' in Figure~\ref{fig:eth_vs_f1}), 
while on `ch-en' datasets the feature is always turned off (corresponding to `chen\_news' and `chen\_oral' in Figure~\ref{fig:eth_vs_f1}).

\begin{figure}[t]
	\centering
	\includegraphics[scale=0.65]{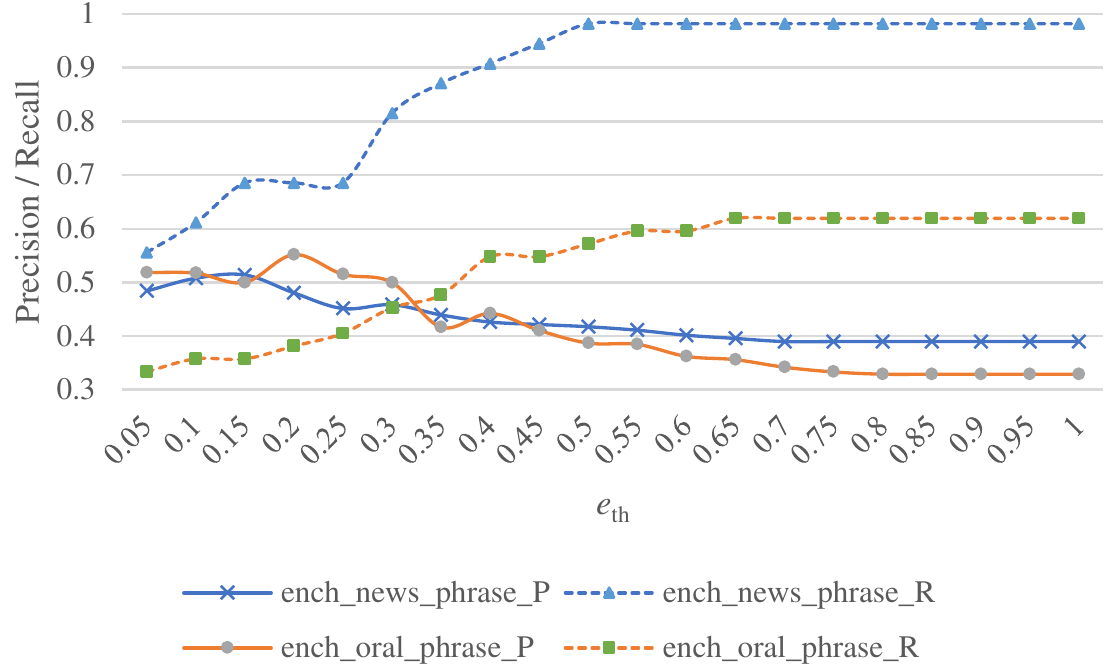}
	\caption{Trends of precisions and recalls of the detection results on `en-ch' datasets with different $e_{th}$ values}
	\label{fig:eth_vs_pr}
\vspace{-4ex}
\end{figure}

As shown in Figure~\ref{fig:eth_vs_f1}, the trends of F-measures on the detection results of different datasets are very similar under the same setting.
Such result indicates the commonality of under-translation detection in various scenarios.
However, differences still exist.
As shown in Figure~\ref{fig:eth_vs_f1}, F-measures on the detection results of `ench\_news', `ench\_news\_phrase', and `chen\_news' peak when $e_{th}$ is around $0.15\sim0.2$.
In contrast, for `ench\_oral' and `chen\_oral', F-measures peak when $e_{th}$ is around $0.2\sim0.25$, and for `ench\_oral\_phrase' the peak value is even higher.
This finding is expected: expressions in news reports tend to be more rigorous than those in casual conversations, 
limiting the scope of word/phrase meanings and thus making it easier to achieve precise translations.
Such finding also shows that our algorithm can be easily tuned on different scenarios.
In addition, the algorithm effectiveness is worse at $e_{th} = 1$ for all curves, where {error rate filtering} is not in use.
Such finding indicates that {error rate filtering} contributes to the algorithm effectiveness.

However, as shown in Figure~\ref{fig:eth_vs_f1}, the influence on F-measures of the detection results is much smaller when phrase identification is on, although the overall F-measures are higher than those in the situation where phrase identification is off.
A potential explanation is that disabling phrase identification causes more false positives on words that actually belong to phrases, given that words can have totally different meanings in phrases.
Filtering the detection results based on error rates helps reduce these false positives and consequently improve the F-measures.
Such finding indicates that phrase identification could actually contribute to the algorithm effectiveness.
To understand why the F-measures change insignificantly, in Figure~\ref{fig:eth_vs_pr}, we show the trends of precisions and recalls with different $e_{th}$ values for `en-ch' datasets when phrase identification is enabled.
As can be seen from Figure~\ref{fig:eth_vs_pr}, as $e_{th}$ goes up, precisions gradually become lower while recalls steadily increase.
The trend is intuitive: less strict filters allow more true positives that contribute to higher recalls and miss more false positives, which lead to lower precisions.
Such finding still shows that error-rate filtering is useful: one of our goals for detecting violations in translations is to build in-house test suites for regression testing and manual inspection.
Keeping test suites small and precise saves time and manual efforts from running or inspecting false positives.


\subsection{Experience of Deployment}
\label{sec:deploy}

\begin{figure}[t]
	\centering
	\includegraphics[scale=0.3]{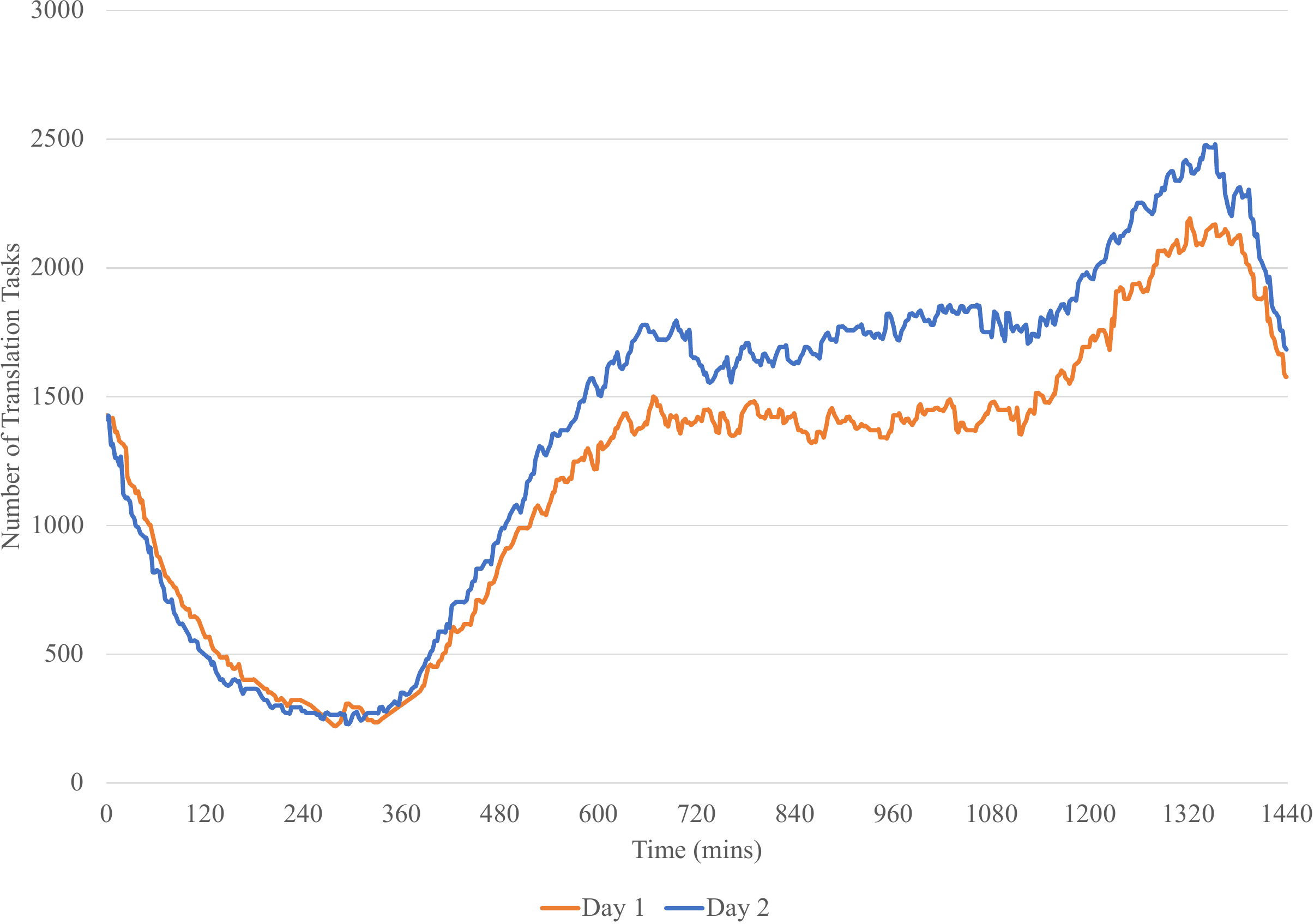}
	\caption{Statistics of unique English-Chinese translations with translation-property violations identified by our algorithms}
	\label{fig:prod_monitor_prob}
\vspace{-2ex}
\end{figure}

\begin{figure}[t]
	\centering
	\includegraphics[scale=0.55]{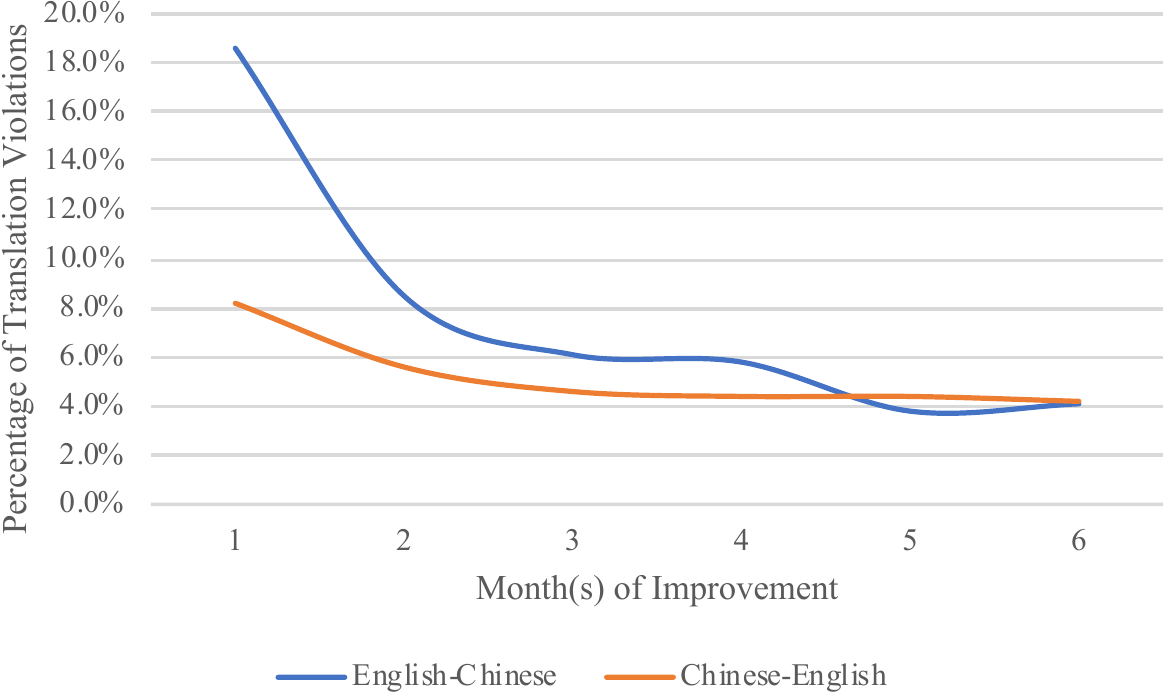}
	\caption{Combined statistics of under- and over-translation violations in real-world English and Chinese translations by our NMT model}
	\label{fig:utrans_pcnt}
\vspace{-4ex}
\end{figure}

By using our approach for in-vivo testing in the production environment, the developers are able to collect translation tasks with unseen types or instances of translation failures and observe the performance of deployed models in real-world usages. 
The developers are also able to handle the translation failures instantly through switching to backup models (\eg, SMT models), improving the overall translation quality nearly effortlessly. 
Figure~\ref{fig:prod_monitor_prob} shows an example 2-day statistics of unique English-Chinese translations in which our algorithms identify any translation-property violation, where Y-axis corresponds to the statistics within each period of 5 minutes.
As can be seen, our approach is able to find over 8 translations with translation-property violations each second during the busiest periods.
On top of that, our algorithms are able to process about 12 million unique translation tasks every day, with over 200 translations processed each second during the busiest periods.
Such result indicates the good performance and applicability of our algorithms and approaches on real-world tasks.
The statistics also show that these translation-property violations are non-trivial in the real world, and in-vivo testing on NMT systems is of great significance.

In addition, not limited to only in-vivo testing, our approach also helps enhance in-house black-box system testing during the NMT-system development. 
By using our approach, the developers manage to find outputs from the NMT system (\ie, the translations) that contain translation failures and are previously unable to be processed by the special test oracle based on the translation quality score (due to missing reference translations). 
Our result also shows the violations decrease significantly after the deployment of our tool.
Such result suggests that the developers are able to quickly locate and diagnose translation failures when test cases fail, based on the information provided by our approach.
Figure~\ref{fig:utrans_pcnt} shows the combined statistics of under- and over-translation violations in real-world English and Chinese translations by our NMT model within 6 months after the deployment of our approach.
As can be seen, the percentage of translations with under- or over-translation violations decreases significantly over time.
More specifically, for English-Chinese translations, the percentage drops from 18.6\% to 4.1\%, while for Chinese-English translations, the percentage drops from 8.2\% to 4.2\%.
Such statistics reflect the effectiveness of our approach in helping development and improvement of a machine translation system.

\subsection{Large-scale Test Suite Generation and Test Results}
\label{sec:appl_rw_tasks}

Our algorithms also help collect 130,000 English and 180,000 Chinese meaningful words/phrases 
(\ie words/phrases with low error rates by learning from training data). 
These words are used as in-house test cases for testing and improving \WeChat{}'s  continuously-improved machine translation model. 
Using our approach not only helps diagnose issues in \WeChat{}'s NMT system but also helps diagnose issues in other competing machine translation systems released by other providers. 
We show some examples of translation-property violations in English and Chinese translations provided by systems from various providers in Table~\ref{tbl:cmpt_mt_issue_ex}.
By analyzing the translations, we gain some insights on potential causes of those violations, showing that our approach is able to detect potential defects lying in the design, implementation, or training data in machine translation systems, with examples as below.

\begin{table}[t]
\centering
\small
\caption{Example issues found in machine translation systems by various providers}
\label{tbl:cmpt_mt_issue_ex}
\begin{tabular}{|p{1.1cm}|p{1.5cm}|p{2.8cm}|p{1.5cm}|}
\hline
Provider\qquad Name	& Original\qquad Text		& Given\qquad\qquad Translation	&	Expected Translation	\\
\hline
Prvd. A & \chstext{成人}		& mature people		&	adult				\\
Prvd. A & \chstext{太好了}	& what fun			&	great				\\
Prvd. B & large-scale		& large-scale		&	\chstext{大规模}		\\
Prvd. B & long-term			& long-term			&	\chstext{长期}		\\
Prvd. B & U.S.				& U.S.				&	\chstext{美国}		\\
Prvd. C & \chstext{蛋糕}		& Runeberg torte	&	cake				\\
Prvd. C & \chstext{酸奶}		& Viili				&	yoghurt				\\
Prvd. D & \chstext{疟原虫}	& p.				&	plasmodium			\\
Prvd. D & \chstext{酶原}		& The original enzyme	&	zymogen		\\
\hline
\end{tabular}
\vspace{-4ex}
\end{table}

For \textit{Provider A}, translations in both examples seem to be too ``straightforward'', \ie, two Chinese words are translated character by character instead of being as a whole. There might be an issue in the design of the model, causing the model to disregard some context information.

For \textit{Provider B}, three English phrases with dots or hyphens remain untranslated. It might also be the problem of model implementation that causes the model to disregard such phrases.

For \textit{Providers C and D}, translations for common food names and terminologies are too specific (\eg, \textit{Runeberg torte} is one type of cakes) or inaccurate (\eg, \textit{p.} is the abbreviation of \textit{plasmodium}, but there are too many words beginning with the letter p). Both providers might need to investigate into their training data.


\section{Related Work} 
\label{sec:related_work}

Tu et al.~\cite{tu2016modeling} propose coverage-based NMT models, which keep track of the attention history by maintaining coverage vectors.
The coverage mechanism assists attention adjustment to provide more chances for untranslated source words.
Both under- and over-translation issues are addressed by the coverage mechanism.
The coverage mechanism is also adopted by Google NMT models~\cite{wu2016google}.
However, such coverage mechanism only alleviates (but cannot eliminate or detect) under-translation and over-translation issues from the perspective of NMT model design. 
It does not reveal defects lying in the implementation or the training data of the model.
Our approach is applicable on general machine translations, aiming to provide common-issue detection regardless of specific translation models, and such common-issue detection could be useful for model development and improvement. In other words, even when coverage-based NMT models are used, our approach is still needed and applicable to detect remaining issues. In addition, our approach can be used even when other NMT models (not being coverage-based ones) are used.  

Our detection algorithms make use of the Item-based Collaborative Filtering~\cite{sarwar2001item} algorithm to construct the word/phrase translation dictionary.
Linden et al.~\cite{linden2003amazon} describe how to use the algorithm to recommend products to Amazon customers, and 
Davidson et al.~\cite{davidson2010youtube} use a similar approach to recommend videos to YouTube users.
We show that such algorithm is also suitable for recommending words/phrases in the scenario of translation quality assurance. 
It is also possible to apply more advanced recommendation algorithms, such as those summarized by Adomavicius and Tuzhilin~\cite{adomavicius2005toward}.


\section{Conclusion} 
\label{sec:conclusion}

In this paper, we present a novel approach for in-vivo testing of an NMT system.
Our approach automatically identifies translation failures without requiring reference translations. 
Our approach focuses on properties of natural language translation that can be checked systematically and uses information from both the test inputs (\ie, the texts to be translated) and the test outputs (\ie, the translations under inspection) of the NMT system. 
Our evaluation conducted on real-world datasets shows that our approach can effectively detect targeted property violations as translation failures. 
Our experiences on deploying our approach in both production and development environments of \WeChat, a messenger app with over one billion monthly active users, demonstrate high effectiveness of our approach along with high industry impact.

\bibliographystyle{IEEEtran}
\bibliography{contents/bibliography}

\end{document}